\documentclass[10pt,twocolumn,letterpaper]{article}

\usepackage{iccv}
\usepackage{times}
\usepackage{epsfig}
\usepackage{graphicx}
\usepackage{amsmath}
\usepackage{amssymb}

\usepackage{adjustbox}
\usepackage{bm}
\usepackage{algorithm}
\usepackage{algpseudocode}
\usepackage{appendix}

\usepackage[breaklinks=true,bookmarks=false]{hyperref}

\iccvfinalcopy %

\def\iccvPaperID{6352} 

\ificcvfinal\fi

\newtheorem{thm}{Theorem}

\newtheorem{axiom}[thm]{Axiom}

\begin{document}

\title{XRAI: Better Attributions Through Regions}

\author{Andrei Kapishnikov\thanks{These two authors contributed equally.},
Tolga Bolukbasi\footnotemark[1],
Fernanda Vi\'egas, 
Michael Terry\\
Google Research\\
Cambridge, MA\\
{\tt\small \{kapishnikov, tolgab, viegas, michaelterry\}@google.com}
}

\maketitle
\ificcvfinal\thispagestyle{empty}\fi

\begin{abstract}
    Saliency methods can aid understanding of deep neural networks. Recent years have witnessed many improvements to saliency methods, as well as new ways for evaluating them. In this paper, we 1) present a novel region-based attribution method, XRAI, that builds upon integrated gradients \cite{sundararajan_axiomatic_2017}, 2) introduce evaluation methods for empirically assessing the quality of image-based saliency maps (Performance Information Curves (PICs)), and 3) contribute an axiom-based sanity check for attribution methods. Through empirical experiments and example results, we show that XRAI produces better results than other saliency methods for common models and the ImageNet dataset.

\end{abstract}

\section{Introduction}

Saliency methods link a deep neural network's (DNN) prediction to the inputs that most influence that prediction. These capabilities can be useful in a wide range of contexts, including debugging a model's prediction, verifying that the model is not learning spurious correlations \cite{ribeiro_why_2016}, and inspecting the model for issues related to fairness \cite{selvaraju_grad-cam_2016}. In this paper, we focus on image-based saliency methods.

A rich set of image-based saliency methods have been developed over the years (e.g., \cite{springenberg_striving_2014, zeiler_visualizing_2013, kindermans_reliability_2017, sundararajan_axiomatic_2017, bach_pixel-wise_2015, chattopadhyay2019neural}).
One common approach of determining salient inputs is to rely on the changes in the model output, such as gradients of the output with respect to the input features. For example, Integrated Gradients (IG) determines the salient inputs by gradually varying the network input from a baseline to the original input and aggregating the gradients \cite{sundararajan_axiomatic_2017}. While existing saliency methods provide very compelling results, there are opportunities to further improve identification of the most important inputs leading to a model's prediction.

Given the potential utility of saliency methods, recent research has begun to critically examine these techniques and has proposed various methods for evaluating them. These evaluation methods provide ways to validate the saliency method's outputs (e.g., to ensure they can be relied upon to explain model behavior) \cite{ancona_towards_2017, adebayo_sanity_2018}, or to empirically measure the methods' outputs, enabling comparison of two or more techniques. For example, ``sanity checks'' have been developed that help determine whether a saliency method's results meaningfully correspond to a model's learned parameters \cite{adebayo_sanity_2018}, while Sensitivity-n \cite{ancona_towards_2017} empirically measures the quality of a saliency method's output by comparing the change in the output prediction to the sum of attributions. 

\begin{figure}[H]
    \centering
    \includegraphics[width=0.99\columnwidth]{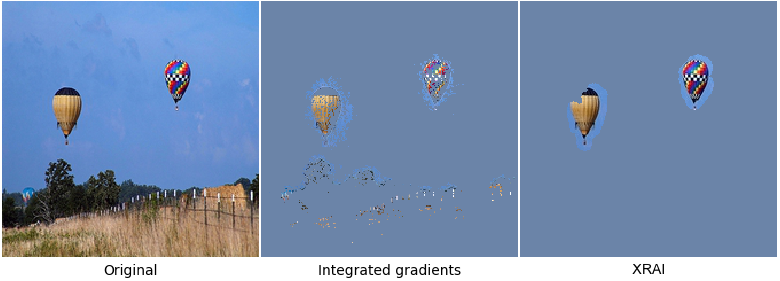}
    \caption{Comparison of Integrated Gradients (middle) and the proposed XRAI method (right) at 5\% area threshold for object class ``balloon''. Areas that may have high pixel-level attributions are removed at the region level as they sum up close to zero.}
    \label{fig:intro}
\end{figure}

\begin{figure*}[!th]
    \centering
    \includegraphics[width=0.99\textwidth]{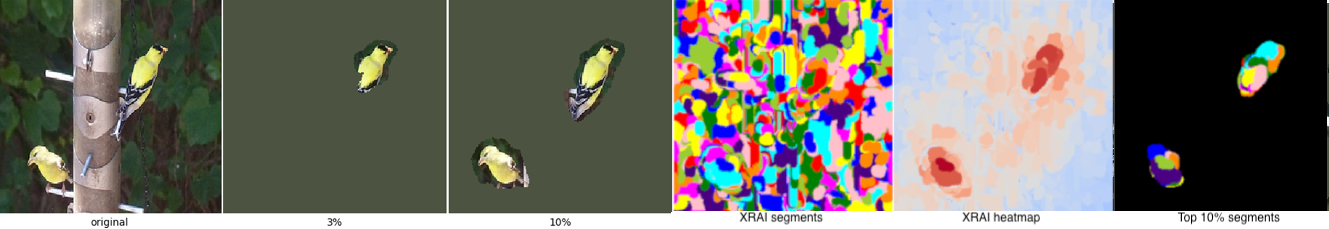}
    \caption{XRAI is a saliency method that incrementally grows attribution \emph{regions}. In this figure, the most important regions for classifying the image as ``goldfinch'' are revealed for different area thresholds: the technique first identifies one bird (3\%), then two birds (10\%). The pool of possible segments to choose from are represented by the colored regions (XRAI segments). A saliency heatmap indicates which of these regions provide the most predictive power (XRAI heatmap). Finally, the most salient segments at area threshold of 10\% (according to XRAI's ranking) are shown. Notice that XRAI constructs the bird from numerous segments and does not solely rely on segment quality.}
    \label{fig:ent}
\end{figure*}

In this paper, we make three sets of contributions. 
First, we propose a novel region-based saliency method, XRAI (Figure \ref{fig:ent}), based on the widely used Integrated Gradients (IG) \cite{sundararajan_axiomatic_2017}. 
Our method first over-segments the image, then iteratively tests the importance of each region, coalescing smaller regions into larger segments based on attribution scores. Through examples and empirical results, we show that this strategy yields high quality, tightly bounded saliency regions that outperform existing saliency techniques. Importantly, XRAI can be used with any DNN-based model as long as there is a way to cluster the input features into segments through some similarity metric (e.g. color similarity in images).

Second, we add to the growing body of sanity checks for attribution methods \cite{adebayo_sanity_2018} by introducing a perturbation-based sanity check that can be used to test the reliability of an attribution method. Our sanity check can be seen as a relaxed version of the Sensitivity-N measure \cite{ancona_towards_2017}, where we require features that cause non-zero change in the output to have at least non-zero attributions. In applying this sanity check, we found that Gradients \cite{baehrens_how_2010, simonyan_deep_2013}, Gradients*Input, and Integrated Gradients can sometimes fail it, even for very sensitive features that cause the prediction to completely change with minimal modification. A key insight from this sanity check is that compared to pixel-level IG attributions, region-level sums of pixel-attributions are more robust (Figure \ref{fig:intro}).

Finally, we introduce a pair of evaluation metrics for empirically assessing the quality of saliency techniques: Accuracy Information Curves (AICs) and Softmax Information Curves (SICs), both similar in spirit to receiver operating characteristics displays (ROC).
These measurement methods are inspired by the bokeh effect in photography, which consists of focusing on objects of interest while keeping the rest of the image blurred. In a similar fashion, we start with a completely blurred image and gradually sharpen the image areas that are deemed important by a given saliency method. Gradually sharpening the image areas increases the information content of the image. We then compare the saliency methods by measuring the approximate image entropy (e.g., compressed image size) and the model's performance (e.g., model accuracy). Collectively, we dub these metrics Performance Information Curves (PICs). We validate these techniques by showing how rankings of popular saliency methods' outputs align with visual results.

In sum, this paper makes the following contributions:
\begin{itemize}
    \item We introduce XRAI\footnote{See https://github.com/PAIR-code/saliency for the implementation and more details.}, a novel region-based attribution method. XRAI can be applied to any DNN model, yields empirically superior results to other methods, and achieves performance that is competitive with other techniques.
    \item We introduce a  perturbation-based sanity check for testing saliency methods. %
    \item We introduce two metrics for measuring the quality of attribution methods, Accuracy Information Curves and Softmax Information Curves. We demonstrate their utility and validity by applying them to popular saliency methods run on the ImageNet dataset with Inception \cite{szegedy_going_2014} and Resnet50 \cite{he_deep_2015} models as well as comparing them with the standard localization metrics.
\end{itemize}

\begin{figure*}[!th]
    \centering
    \includegraphics[width=0.93\textwidth]{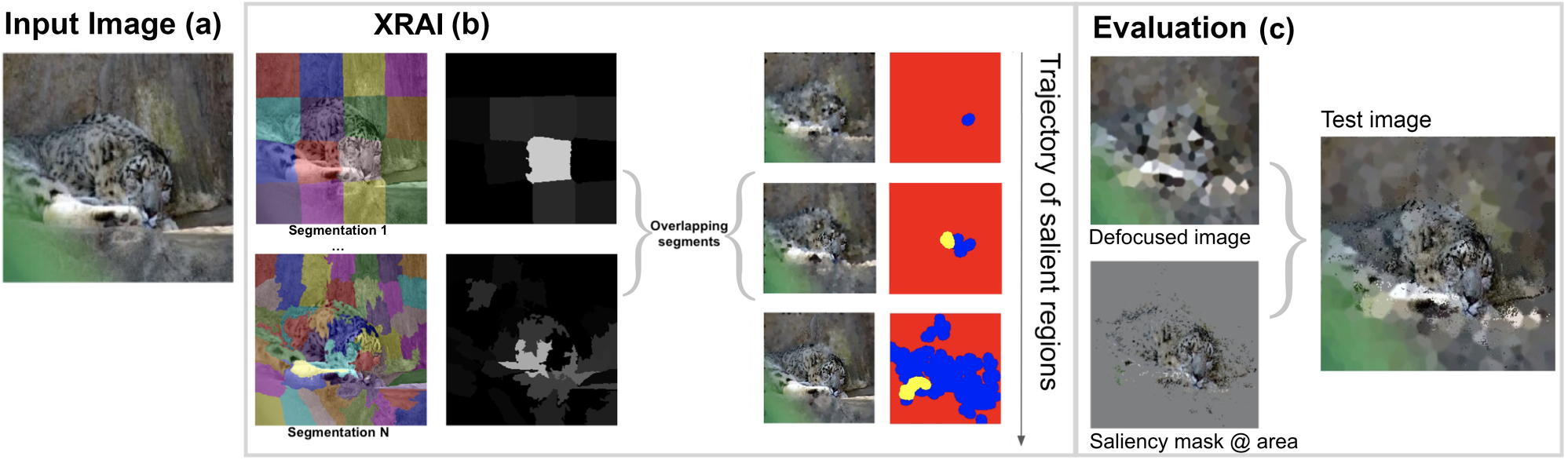}
    \caption{\textbf{(a)} Input leopard image. \textbf{(b)} XRAI's segmentation process. First, the image is over-segmented to many overlapping regions of various shapes, then the segments are gradually added with respect to their integrated gradients density. The ranking of region importance can be recovered from the trajectory. In this case, XRAI reconstructed the leopard's face, yielding a correct classification, then added the body and rest of the image. \textbf{(c)} Diagram of the evaluation method for a single image and a given area threshold. The unfocused image and salient region mask get combined to produce the saliency-focused image. This image is fed back to the classifier to measure performance. \label{fig:sys_diag_main}}
    \vspace{-0.3cm}
\end{figure*}

The rest of the paper is structured as follows. We first review related work, then introduce a perturbation-based sanity check. We then describe XRAI and its implementation. We introduce two evaluation methods, Accuracy Information Curves and Softmax Information Curves to complement existing evaluation metrics for saliency methods. We then present results from a series of experiments that compare XRAI with GradCAM, Gradient, Gradient*Input, and IG with different baselines, and show representative samples of each technique on the ImageNet dataset \cite{russakovsky_imagenet_2014}. The overall diagram of our evaluation and attribution methods can be found in Figure \ref{fig:sys_diag_main}.

\section{Related Work}

Numerous methods of attributing inputs to output predictions have been proposed.
One set of methods modify the inputs and measure the effect this perturbation has on the output by performing a forward pass through the network with these modified inputs \cite{fong_interpretable_2017, carter_what_2018}.
As an example, \cite{zeiler_visualizing_2013} proposed a method to visualize the prediction score of the correct class as a function of a gray patch occluding the original image. LIME \cite{ribeiro_why_2016} fits a simpler local model to approximate the prediction surface by querying the model in a close neighborhood of the original input.

While perturbation based-methods allow one to directly estimate the impact of a feature subset on the output, they require multiple queries to the model, making them slow \cite{zintgraf_visualizing_2017}. Moreover, performance degrades as a function of the number of features. Finally, the nonlinear nature of neural networks means that the results are only reliable for the exact subset and modification of the features, making it challenging to obtain a reliable estimate for all perturbations (i.e., there is exponential complexity if one tries every subset of features).

A second set of approaches calculate attributions by back-propagating the prediction score through each layer of the network, back to the input features. These methods are in general faster than perturbation-based methods since they usually require a single or constant number of queries to the neural network (independent of the number of input features). Some examples include Guided Backprop \cite{springenberg_striving_2014}, DeConvNet \cite{zeiler_visualizing_2013}, Integrated Gradients \cite{sundararajan_axiomatic_2017}, Layer-wise Relevance Propagation \cite{bach_pixel-wise_2015}, SmoothGrad \cite{smilkov_smoothgrad:_2017}, Deep-Lift \cite{shrikumar_learning_2017}, GradCAM \cite{selvaraju_grad-cam_2016}, Gradients*Input \cite{shrikumar_not_2016}, and others \cite{kindermans_learning_2017, montavon_explaining_2017}. The method proposed in this paper, XRAI, falls within this family of attribution methods.

Recently, the reliability and validity of saliency methods have been critically examined. In particular, researchers have discovered that many saliency maps are fragile against adversarial attacks \cite{ghorbani_interpretation_2018}. Adebayo et al. \cite{adebayo_sanity_2018} introduced a set of sanity checks for saliency methods to ensure they produce valid results. Specifically, they introduced two general classes of sanity checks: similarity metrics and visualizations. In applying these checks, they found that some saliency methods produce similar outputs regardless of whether a model is trained or completely random. This result indicates that some saliency methods' results cannot be considered completely reliable explanations for a trained model's behavior. In the context of this paper's research, they found that Integrated Gradients did not convincingly pass the visualization sanity check. As we show later, our technique introduces some key modifications to Integrated Gradients that enables it to successfully pass this sanity check.

Our measurement methods are related to the smallest sufficient region, where the goal is to find the smallest region such that the prediction is the correct class \cite{dabkowski_real_2017}. However, we do not constrain the measurement to the smallest region and formulate our metrics as importance ranking of all regions in the image with respect to their attribution. Although the smallest sufficient region gives a high level insight, it fails to capture less important but potentially useful or problematic regions in the image.

\section{Perturbation Sanity Check}
\label{sec:sanity_check}

In this section, we propose an axiom that functions as a sanity check for the attribution methods. While this axiom motivated XRAI (described in the next section), it is applicable to any attribution method.

\begin{axiom}
Perturbation-$\epsilon$: Given $\epsilon$, for every feature $x_i$ in an input $\bm{x} = [x_1, ..., x_N]$ where all features except for $x_i$ are fixed, if the removal (setting $x_i=0$) of feature $x_i$ causes the output to change by $\Delta y$, then Perturbation-$\epsilon$ is satisfied if the inequality $attr(x_i) \geq \epsilon * \Delta y$ is satisfied.
\end{axiom}%

This axiom can be seen as a relaxation of Sensitivity-1 from \cite{ancona_towards_2017}. More specifically, the right and left sides of the inequality are exactly equal to each other for $\epsilon=1$ when Sensitivity-1 is satisfied. Note that even for cases when Sensitivity-1 is not satisfied, this axiom should be satisfied for a large enough $0< \epsilon \leq 1$ since this implies that the features that change the output after removal should have non-zero attributions.

We propose a simple simulation to test this axiom. A neural network ideally learns to approximate a function $f(\bm{x})$ for any input $\bm{x}$. We can directly sample $f$ with some properties to test the axiom. For the sake of simplicity, assume two pixel images where the input features are $x_1$ and $x_2$. We define the function as:
\begin{equation}\label{eqn:simulation}
    f(x_1, x_2) = \begin{cases}
              1.0 \quad if \: x_1 = 127 \: and \: x_2 = 127 \\
              uniform \sim [0.0, 0.5] \quad if \: x_1, x_2 \in grid
            \end{cases} 
\end{equation}

\begin{figure*}[!th]
    \centering
    \includegraphics[width=0.4\textwidth]{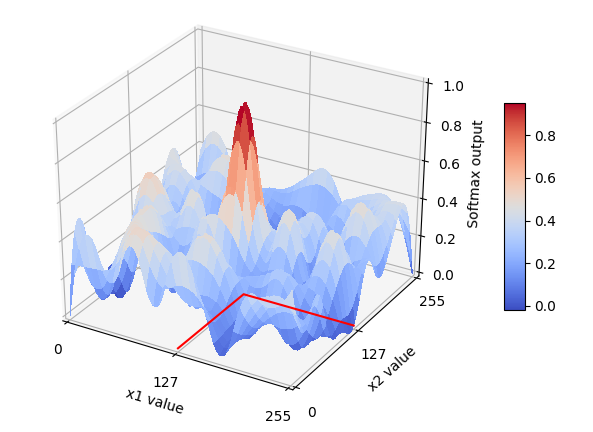}
    \includegraphics[width=0.4\textwidth]{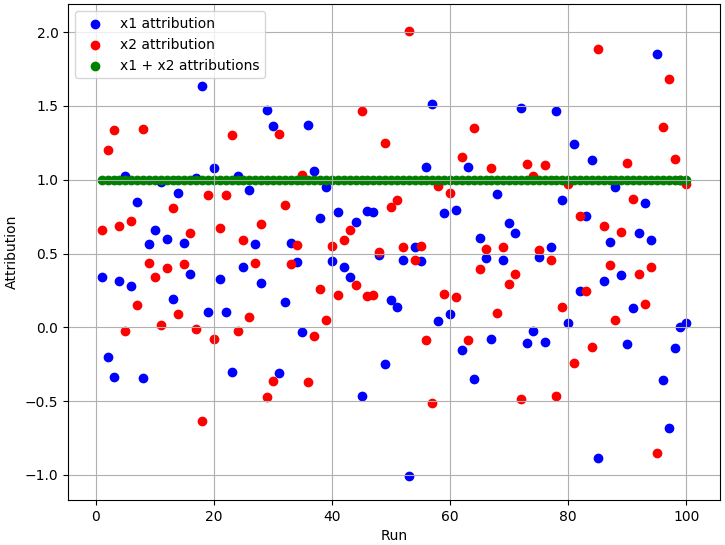}
    \caption{IG pixel-level attributions are not reliable. Although any deviation from the input $x_1=127, x_2=127$ has significant impact on the output score, integrated gradients may assign 0 attribution to either one of the pixels.}
    \label{fig:pixel_attr_fun}
    \vspace{-0.3cm}
\end{figure*}

We use bi-cubic interpolation to obtain a smooth continuous function from a 20x20 grid in the range [0, 255]. One instance of this function is illustrated in Figure \ref{fig:pixel_attr_fun}. By construction, this function peaks at (127, 127) and any small change in $x_1$ or $x_2$ will sharply drop the classification score. Specifically, any point on the grid except for (127, 127) is less than $0.5$, flipping the prediction to negative. For an attribution method to satisfy Perturbation-$\epsilon$ for an $\epsilon \approx 0$,
it needs to attribute a small non-zero value to both $x_1$ and $x_2$ for all functions of the form defined in Equation \ref{eqn:simulation}.
Our sanity check is then to sample many of these functions and test the saliency method for each run. The saliency method fails the test if there are some instances of this function where the attribution of one or both of the input features is zero (or $\epsilon$ close to zero, from the axiom). 

We now show that some of the popular pixel-level attribution methods do not pass this simple sanity check. First, the gradient with respect to inputs at the peak is zero. Therefore, it is easy to show that Gradient \cite{simonyan_deep_2013} and Gradient*Input \cite{shrikumar_not_2016} do not pass, and attribute %
zero to all features. Even in the case of real networks, the gradient around the peak can be non-zero but point to an arbitrary direction, failing the test with non-zero probability that grows with $\epsilon$. It has been previously observed that even when there isn't a peak, gradients fail at saturated areas \cite{sundararajan_axiomatic_2017}. Integrated Gradients (IG) tackles this issue by using a baseline image and computing the gradients along a path. By construction, IG satisfies the Completeness property, which guarantees that at least some features get non-zero attribution in the case when the softmax value is non-zero.

Figure \ref{fig:pixel_attr_fun} shows the attribution values of $x_1$ and $x_2$ for Integrated Gradients. Surprisingly, Integrated Gradients can have zero attribution for $x_1$ or $x_2$ in some cases, even when the input is at the peak. This is problematic, because the attribution of a very important pixel is heavily influenced by the behavior of a neural network in regions that are far away and not important for the classification around the peak. We expect some degree of locality where far away behavior should not affect the attributions for a particular input.

It is important to emphasize that for all generated functions, both $x_1$ and $x_2$ are equally important around the peak. The difference between the runs is due to the different model behavior outside the "true label" classification region. In practice, these differences in behavior can be caused by different model weight initializations or learning rates. A good attribution method should be immune to these factors.

\begin{figure}[!ht]
    \centering
    \includegraphics[width=0.90\columnwidth]{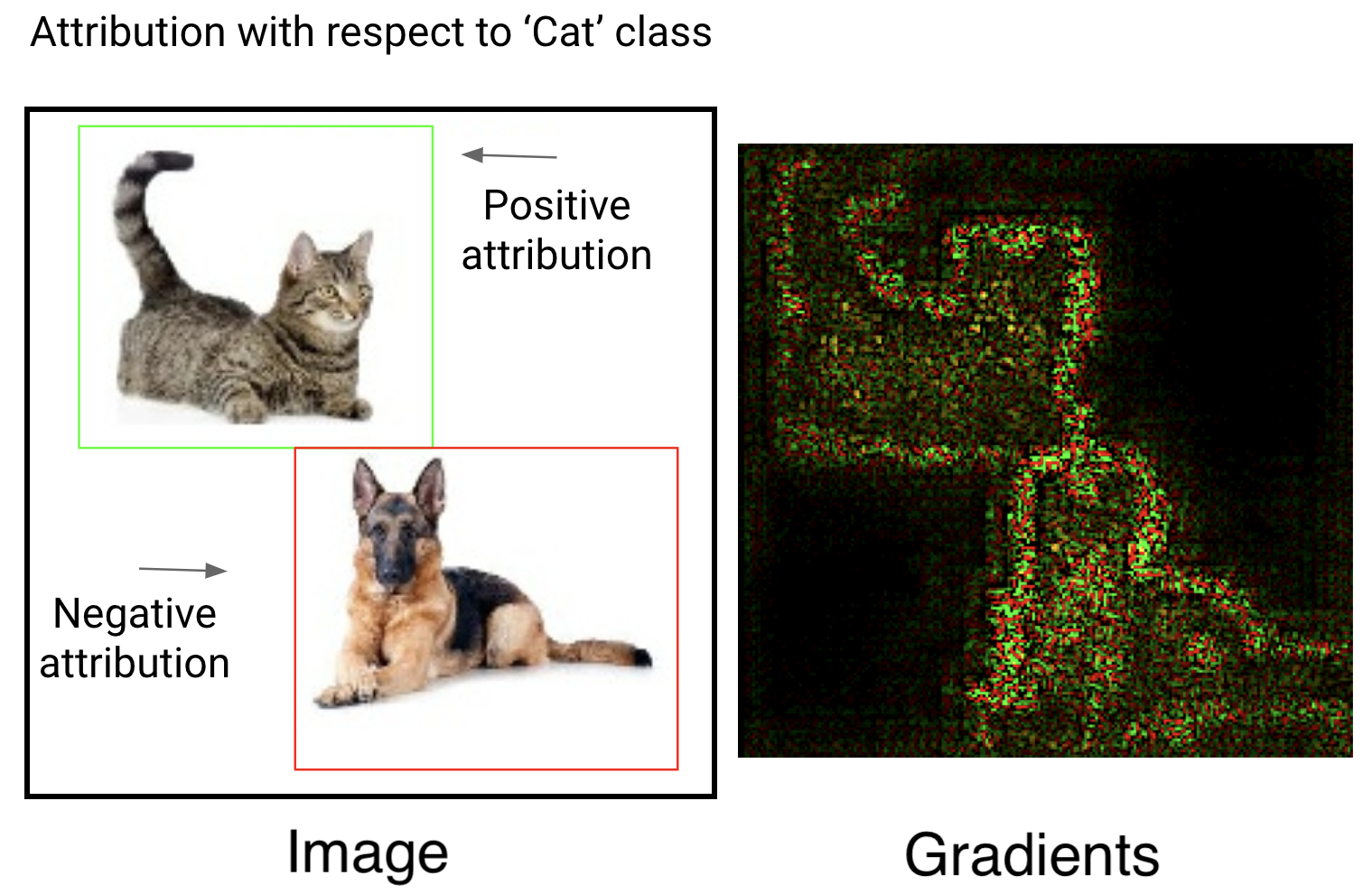}
    \caption{\textbf{(Left)} A single image that contains both ``cat'' and ``dog'' and attributions calculated w.r.t. object class ``cat''. Although there are negative and positive pixel level attributions, the overall sum of IG attributions within the bounding box of the ``cat'' is positive and the bounding box of the ``dog'' is negative. XRAI uses that information to find truly salient areas. \textbf{(Right)} Gradients taken for the image with respect to the  ``cat'' class. Regardless of the selected class, gradients act as an edge detector and attribute both objects.}
    \label{fig:region_attr_cat_dog}
    \vspace{-0.3cm}
\end{figure}

We observed that although individual pixels can be unreliable for Integrated Gradients, larger regions that cover the objects of interest have more reliable attributions. Figure \ref{fig:region_attr_cat_dog} shows an example of this phenomenon for an image with a cat and a dog. Based on this observation, we propose a region-based method based on a modified version of Integrated Gradients.

\section{XRAI}
\label{sec:xrai}
In this section we first describe the XRAI algorithm, then explain its behavior and validate its reliability. The high-level steps to compute XRAI are shown in Algorithm \ref{alg:xrai}.

\begin{algorithm}
    \caption{XRAI}
    \label{alg:xrai}
    \begin{algorithmic}[1] %
        \State Given image $\bm{I}$, model $f$ and attribution method $g$
        \State Over-segment $\bm{I}$ to segments $s \in S$
        \State Get attribution map $A = g(f, I)$
        \State Let saliency mask $M=\bm{0}$, trajectory $T = []$
        \While{$S \neq \emptyset \:\: and \:\: area(M) <  area(I)$}
            \For{$s \in S$}
                \State Compute gain\footnotemark : $g_s = \sum_{i \in s \setminus M} \frac{A_i}{area(s \setminus M )}$
            \EndFor
            \State $\hat{s} = \arg \max_s g_s$
            \State $S = S \setminus \hat{s}$
            \State $M = M \cup \hat{s}$
            \State Add M to list T 
        \EndWhile
        \State \textbf{return} $T$
    \end{algorithmic}
\end{algorithm}

\footnotetext{Both $s \setminus M$ and $s \cup M$ are possible as gain functions. We report results on union, but the subtraction can produce better results; see the open source implementation for more details.}

\textbf{Segmentation} We use Felzenswalb's graph-based method in the skimage python package \cite{felzenszwalb_efficient_2004} for segmentation. Segmentation methods usually have multiple sets of parameters that change the number and the shape of the segments. We do not want the attribution results to depend on a particular set of hyper-parameters or the quality of the segmentation method. For this reason, we segment the image multiple times using different parameter sets. More specifically, we use a scale parameter within the set [50, 100, 150, 250, 500, 1200] and ignore segments smaller than 20 pixels (the scale parameter mainly affects the size of the segments).
For a single parameter, the union of segments yields the entire image. Therefore, the union of all segments yields an area equal to six times the image area, with the result being that individual segments overlap significantly.

Segment boundaries typically align with edges in the image. To derive saliency maps, it is desirable that the segments \emph{include} edges, since attributions on either side of a thin edge are often related to each other. For that purpose, we dilate the segment masks by 5 pixels to obtain our final set of segments.

\textbf{Attribution} For attribution, XRAI uses Integrated Gradients with black and white baselines. This choice is motivated as follows.

\begin{figure}[!ht]
    \centering
    \includegraphics[width=0.45\textwidth]{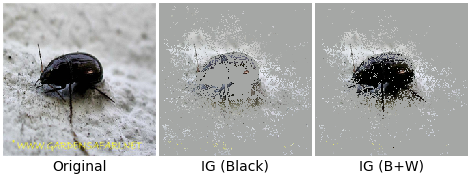}
    \caption{Black baseline IG fails to attribute center of the black beetle, attributing bright background pixels instead (center). The B+W baselines result in attribution of both bright and dark pixels (right).}
    \label{fig:black_beetle}
    \vspace{-0.3cm}
\end{figure}

With the Integrated Gradients technique, using a black image as baseline reduces attribution of dark input pixels. For example, the dark pixels on the beetle in Figure \ref{fig:black_beetle} are not attributed, although they may be more important than the brighter ones. In fact, an RGB value of (0, 0, 0) will receive exactly 0 attribution. This is apparent from the Integrated Gradients' formula:
\begin{equation}
    IG_i(x) = (x_i - x^\prime_i) \int_{\alpha =0}^{1} \frac{\partial F(x^\prime + \alpha \times (x - x^\prime))}{\partial x_i}d\alpha
\end{equation}
where $(x_i - x^\prime_i)$ is the difference between the input pixel $i$ and the corresponding baseline pixel.

More generally, any single baseline integrated gradient will be insensitive to pixels that are equal or close to the baseline image. 
One alternative in practice is to use multiple random baselines. This approach has two drawbacks. First, the resulting saliency maps are not consistent and change at every run due to the randomness of the baseline. Second, this method is more likely to introduce patterns that cause spurious attributions due to random weighing of pixels with respect to their closeness to the random baseline. 

To address these issues, XRAI uses black and white baselines. In this way, the sum of the weight term for any pixel in the image is guaranteed to be $1.0$ since $|x - 1.0| + |x - 0.0| = 1.0 \:\: \forall x \in [0.0, 1.0]$, where $x$ is the input pixel value and $1.0$ and $0.0$ correspond to black and white baselines. Therefore, all pixels get an equal chance of contributing to the attributions regardless of the distance from the baselines. In addition, this method produces consistent saliency maps. %

\textbf{Selecting regions} To select regions, XRAI leverages the fact that IG satisfies Sensitivity-N \cite{ancona_towards_2017}, where the sum of all attributions for an input is equal to the input softmax value minus the baseline softmax value. Given two regions, the one that sums to the more positive value should be more important to the classifier. From this observation, XRAI starts with an empty mask, then selectively adds the regions that yield the maximum gain in the total attributions per area. The algorithm runs until it obtains the full image as the mask or runs out of regions to add. One can see the trajectory of masks as the importance ranking of the regions. %

It has been reported that it is easier for humans to understand regions instead of pixels \cite{sundararajan_exploring_2019}. Methods such as GradCAM have regions with inherent smoothing due to mapping from a lower resolution convolutional layer to the input layer. Although they produce smoother regions due to this effect, it is usually a side effect. XRAI also constructs smooth areas, which typically encapsulate the whole object by merging salient segments. While the segments XRAI produces often correspond to human intuition about what constitutes a semantically meaningful region, there is nothing explicitly encoded in XRAI's algorithm to choose segments that align with how humans perceive images.

\subsection{XRAI: Behavior and validation with sanity checks}
In this section, we describe the features of XRAI that help to create reliable output, as well as how it is more robust against sanity checks described in Section \ref{sec:sanity_check} and Adebayo et al. \cite{adebayo_sanity_2018}.

Gradient-based methods work by measuring the sensitivity of model output with respect to changes in individual input features. This is accomplished by taking the partial derivative $\frac{\partial f}{\partial x_i}$ of input feature $i$. The partial derivatives do not directly reveal whether a particular input feature is contributing to a predicted class or not; they merely show whether the changes in the input change the model prediction. As a result, some of the features may not be relevant to the predicted class but still have high attribution.

An example of this issue can be demonstrated with an image of a cat and dog Figure \ref{fig:region_attr_cat_dog}. Even if a model classifies the image as containing a cat, the derivatives of the dog pixels produce non-zero attributions since removing the dog changes the cat prediction.  This happens because removing the dog reduces the softmax output associated with the dog class.  As a result, the softmax output of the cat increases. Due to this limitation, gradient-based methods can act as edge detectors \cite{adebayo_sanity_2018} by attributing any high contrast regions that change the softmax output of any of the possible classes, including classes different from the prediction.

XRAI addresses this issue by identifying regions that are relevant to the predicted class and discarding irrelevant ones.
It is inspired by the ``Completeness'' properties of IG \cite{sundararajan_axiomatic_2017}, which states that the sum of all attributions is equal to the difference between the model output at the input $x$ and the baseline $x'$. That means that the image regions that truly contribute to the predicted class should have high positive attribution; regions that are unrelated to the prediction should have near-zero attribution; and regions that contain competing classes should have negative attribution. This quality can be demonstrated with Figure \ref{fig:region_attr_cat_dog} as well. If we start with an empty image and add the region with the cat, the output of the model will change from near $0.0$ to $1.0$. Thus, due to the ``Completeness'' properties, the delta of the attributions must be positive. The introduction of the region with the dog causes the output of the model to drop from $1.0$ to $0.6$, thus changing the attributions by $-0.4$. Likewise, if changing the background does not change the model prediction, the attribution of the background is near-zero.

XRAI also addresses the reliability issue of individual pixels (Section \ref{sec:sanity_check}) by combining individual pixels into a group of pixels. As Figure \ref{fig:pixel_attr_fun} shows, even though the attribution of individual pixels $x1$ and $x2$ are unreliable, their combination always adds up to $1.0$.

\begin{figure}[htb]
    \centering
    \includegraphics[width=0.36\columnwidth]{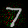}
    \includegraphics[width=0.36\columnwidth]{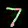}
    \caption{Attributions produced by IG with black+white baselines for a network trained with random labels (left) versus attributions for network trained with correct labels (right). When black+white baselines are used, the random network attributes randomly, whereas the trained network attributes digit pixels. Therefore, IG with black+white baselines passes the sanity check \cite{adebayo_sanity_2018}.}
    \label{fig:visual_sanity_check_mnist}
    \vspace{-0.3cm}
\end{figure}

Adebayo et al. \cite{adebayo_sanity_2018} raise valid concerns about the reliability of saliency methods in general, and propose an image similarity-based sanity check that checks the similarity of the saliency map as they gradually randomize the layers of a neural network. The test fails if the saliency method produces very similar attribution maps for random and trained neural networks, showing that the attributions are not correlated to the trained model. An analysis of IG suggests that its attributions for random networks are uncorrelated with those of trained networks  \cite{sundararajan2018note}. Our method operates on top of IG's outputs, which shows that if IG is dependent on the model, then XRAI will depend on it as well. 
Adebayo et al. also suggest performing a visual analysis for attributions from a random and trained neural network. Figure \ref{fig:visual_sanity_check_mnist} demonstrates that IG with Black and White baselines passes this test.

\section{Assessing Attribution Quality Through Blurring and Iterative Testing}
\label{sec:measure_attr}

Conducting quantitative evaluations of attribution methods is desirable but challenging. %
One of the primary difficulties with saliency methods is that, unlike many machine learning tasks, there is no ground truth for comparison, which makes it difficult to obtain precise numerical results. One common practice to address this challenge is to find the smallest subset of features that yield the correct classification \cite{dabkowski_real_2017}. If the saliency method is in fact identifying pixels important to the model's prediction, this should be reflected in the model's output for the reconstructed image (e.g., the model should predict the same class as in the original image). However, the nonlinear nature and high dimensionality of neural networks also make it difficult to assess the effect of a subset of features on the output in isolation. For example, it has been observed that merely masking the image pixels out of the region of interest causes unintended effects due to the sharp boundary between the masked and salient region \cite{dabkowski_real_2017}. It is therefore crucial to minimize such adversarial effects when testing for the importance of a feature subset.

\begin{figure}[ht]
    \centering
    \includegraphics[width=0.90\columnwidth]{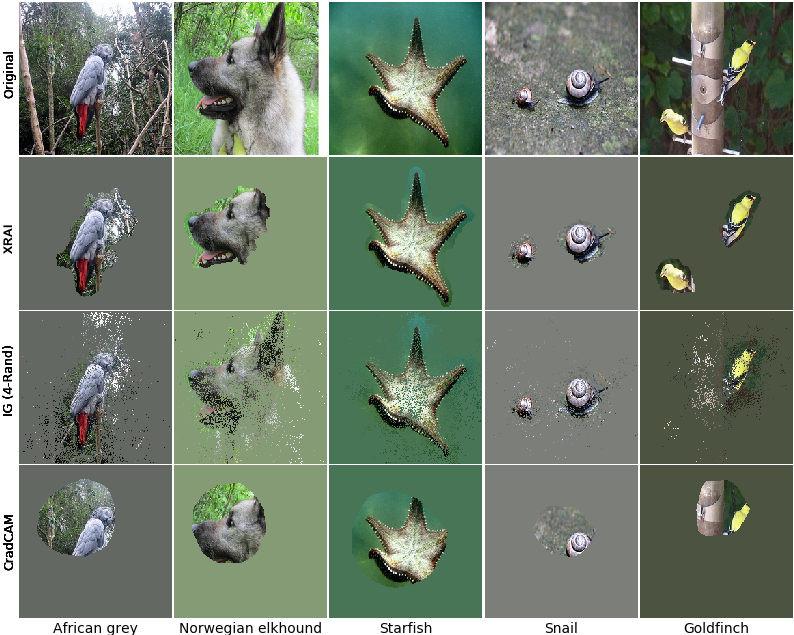}
    \caption{XRAI (2nd row) compared to Integrated Gradients with random baselines (3rd row) and GradCam (bottom row). GradCam can produce blobby regions, whereas XRAI tend to create regions tightly bound around identified objects.}
    \label{fig:gradcam_sig_visual}
    \vspace{-0.3cm}
\end{figure}

Based on these observations, we start with a blurred version of the image (effectively removing information from the image, see Fig. \ref{fig:sys_diag_main}), add back the pixels the saliency method determines are important, measure the entropy of the resulting image, and performing classification on this new image. Model results (e.g., accuracy) are then mapped as a function of the calculated entropy, or \emph{information level}, for each input image (the rationale for using image entropy is expanded below). We call the resultant plots Performance Information Curves (PICs). These plots allow one to more easily compare saliency methods.

The strategy of gradually re-introducing content and monitoring model outputs has the additional advantage of revealing the most important pieces for the model's prediction, the next-to-most important pieces, and so on. For example, imagine an image with a husky in the foreground and snow on the background. Even when the husky is the most important region and leads to correct classification, it is valuable for the practitioner to discover that the background snow is also an important region. In addition, starting with a blurred image instead of directly masking the salient pixels produces more realistic images. As observed in Figure \ref{fig:sys_diag_main}, the resulting images look like real images with the \emph{bokeh} effect. From this point on, we will refer to the blurred images with focused salient regions as \textit{bokeh} images. %

\begin{figure*}[th]
    \centering
    \includegraphics[width=0.24\textwidth]{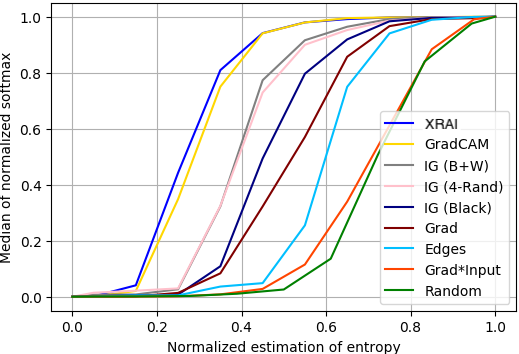}
    \includegraphics[width=0.24\textwidth]{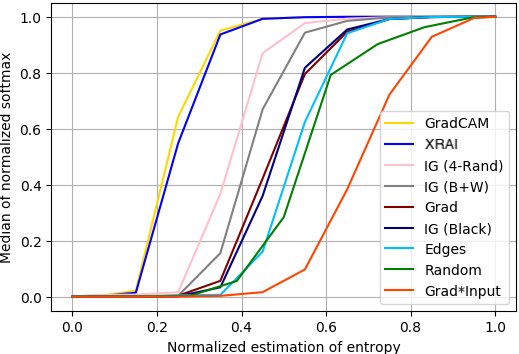}
    \includegraphics[width=0.24\textwidth]{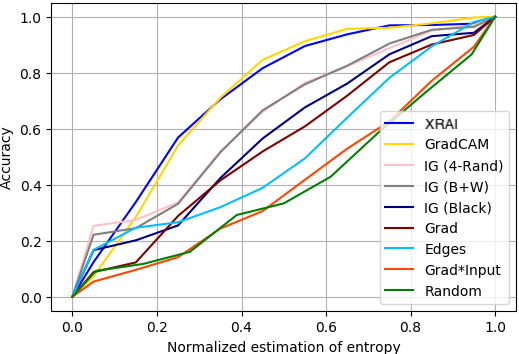}
    \includegraphics[width=0.24\textwidth]{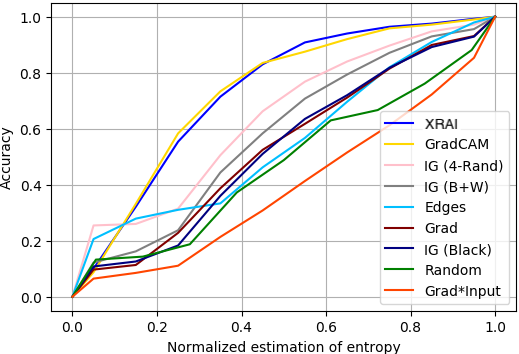}
    \caption{Median softmax information curves (SIC) (Left) and accuracy information curves (AIC) (Right) of various saliency methods for models Inception \cite{szegedy_going_2014} and Resnet50 \cite{he_deep_2015}, respectively. XRAI outperforms all other methods while GradCam follows closely.}
    \label{fig:sic_curve_median}
\end{figure*}

\begin{figure*}[th]
    \centering
    \includegraphics[width=0.99\textwidth]{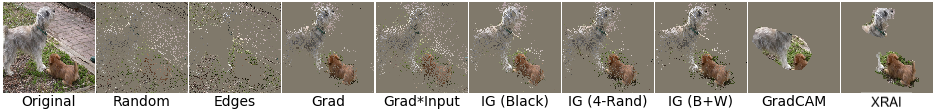}
    \caption{A visual comparison of all methods at a fixed area threshold (20\%). When the image is not totally focused on the object of interest (Norfolk terrier), the edge method chooses many unrelated areas. IG with different baselines produce grainy images. XRAI chooses both dogs and produces the correct prediction when fed back to the image.}
    \label{fig:methods_comp_vis}
    \vspace{-0.3cm}
\end{figure*}

PICs are similar in spirit to ranking methods based on the smallest sufficient region \cite{dabkowski_real_2017}, but provide a more complete view of a saliency method's quality. To produce a PIC, we aggregate the performance per information level over all the samples in a dataset.
The area under PIC, then, is used to measure the performance of a saliency method.
In this sense, our measurement can be seen as analogous to the area under Receiver Operating Characteristics (ROC) curves instead of comparing the threshold at which a model has most accuracy.

We propose two variants of PICs that vary in the measure of performance: \emph{Accuracy Information Curve (AIC)} and \emph{Softmax Information Curve (SIC)}. For AIC, the y-axis is the accuracy calculated over all the images for each bin of image information level. For SIC, the y-axis is the proportion of the original label's softmax for the \textit{bokeh} versus the softmax for the original image.

\noindent \textbf{Using Information Level for Plots} In plotting the quality of a saliency map, one could potentially map the calculated values (accuracy or softmax) as a function of the amount of original content re-introduced in the \textit{bokeh} image (or, more generally, the proportion of the area of the saliency mask to the area of the image). %
However, this approach heavily penalizes saliency methods that produce connected and coherent regions. This is because a grainy saliency map can span a significant area of the image for the same number of pixels. This is counterintuivite from a human understanding standpoint, since it is known that cluttered representations result in higher cognitive load \cite{sundararajan_exploring_2019}. A small number of pixels representing edges can also convey more information than pixels in smooth regions (e.g., a sketch of a cat is easy to recognize compared to a little patch of fur on the cat, even when they may contain the same number of pixels).

To address these issues, we use the entropy as the information content in the image, and plot results as a function of the amount of information in the input image. The compressed size of an image can serve as a useful proxy for the approximate entropy of an image, as it is not feasible to exactly measure it. We use WebP \cite{google_developers_webp_2015}, a popular lossless image compression format, and define the information to be the proportion of the compressed size of the \textit{bokeh} image to the original image.

\section{Experiments}
In this section, we evaluate and compare XRAI to other popular saliency methods on ImageNet validation images \cite{russakovsky_imagenet_2014} using the measuring framework described in Section \ref{sec:measure_attr}, as well as through visual inspection. We show that the rankings of all the methods with respect to our measuring framework align with general consensus, validating the framework's effectiveness.

\subsection{Area under SIC and AIC}

\begin{table}[ht]
	\centering
	\caption{Area under curve for SIC and AIC for all methods.\label{tab:auc}}
	\begin{adjustbox}{width=0.9\linewidth}
    	\begin{tabular}{|l|ll|ll|}
    		\hline
    		Method  & Resnet50-V2  & & Inception & \\
    		 & SIC  & AIC & SIC & AIC \\
    		\hline
    		XRAI & 0.749 & \textbf{0.728} & \textbf{0.720} & \textbf{0.727} \\
    		GradCam & \textbf{0.760} & 0.727 & 0.703 & 0.724 \\
    		IG (B+W) & 0.575 & 0.579 & 0.601 & 0.634 \\
    		IG (4-Rand) & 0.623 & 0.636 & 0.595 & 0.638 \\
    		IG (Black) & 0.515 & 0.527 & 0.530 & 0.576 \\
    		Grad & 0.521 & 0.532 & 0.480 & 0.543 \\
    		Grad*Input & 0.315 & 0.392 & 0.298 & 0.409 \\    	
    		Edges & 0.473 & 0.552 & 0.403 & 0.514 \\
    		Random & 0.445 & 0.473 & 0.278 & 0.401 \\
    		\hline
    	\end{tabular}
    \end{adjustbox}
    \vspace{-0.35cm}
\end{table}

Based on the results in Figure \ref{fig:sic_curve_median} and Table \ref{tab:auc}, we see that our measurement method agrees with what can be visually observed in Figure \ref{fig:methods_comp_vis}. Random saliency performs poorly from the standpoint of prediction at all information levels. This shows that the predictive power is truly affected by which pixels are chosen, and is not coming from the blurry background image.

The random baseline IG and black and white (B+W) IG are significantly better than the black baseline IG. The black and white baseline IG has the advantage of producing deterministic saliency maps as well as running faster than 4-random basline since the runtime is proportional to the number of baselines.

\subsection{Visual analysis}

Figure \ref{fig:methods_comp_vis} shows sample output for a number of popular methods for a fixed area threshold on an image with two dogs. The variants of IG perform relatively well, but create grainy regions. Edges get more background attribution than within the dogs. This image visually demonstrates that the edge method typically only performs well when there is a single object that takes most of the image, as the only edges belong to the true object in that case.

Because of the good performance of GradCAM (as observed empirically and visually), we provide a more detailed comparison of XRAI with GradCAM in Figure \ref{fig:gradcam_sig_visual}. In particular, GradCAM tends to pick one region and gradually expand it as the threshold is increased. In comparison, XRAI can focus on multiple areas. 
This effect is illustrated in Figure \ref{fig:gradcam_sig_visual} with the images of a parrot, dog and starfish where our method covers the object of interest tightly, whereas GradCAM produces smooth circular regions. We also observe that in the presence of multiple objects, GradCAM tends to focus on an area in between the objects of interest. In the case of snails and birds, one can see that the focus is shifted towards the second object when the area threshold is not enough to cover both objects.

\subsection{Weakly Supervised Localization}
We also ran standard localization metrics \cite{cong2018review} on the ImageNet segmentation dataset. Table \ref{tab:localization} shows XRAI outperforming IG and GradCam on all of the metrics. Interestingly, the rankings are similar to our SIC and PIC metrics. These results indicate that our evaluation methods can be used as a proxy to localization tests when ground truth segments are not available for a dataset.

\begin{table}[hbt]
\centering
{\footnotesize
\caption{ImageNet segmentation dataset localization metrics.\label{tab:localization}}
\begin{tabular}{|l|l|l|l|l}
\cline{1-4}
method:  & AUC            & F1             & MAE            &  \\ \cline{1-4}
IG\_B    & 0.710          & 0.674          & 0.219          &  \\ \cline{1-4}
IG\_4RND & 0.709          & 0.674          & 0.223          &  \\ \cline{1-4}
IG\_B+W  & 0.729          & 0.681          & 0.216          &  \\ \cline{1-4}
GradCAM  & 0.742          & 0.715          & 0.194          &  \\ \cline{1-4}
XRAI     & \textbf{0.836} & \textbf{0.786} & \textbf{0.149} &  \\ \cline{1-4}
\end{tabular}}
\vspace{-0.3cm}
\end{table}

\section{Conclusion}

In this paper, we propose a perturbation-based sanity check for saliency maps, a new algorithm that uses region information to improve upon integrated gradients (XRAI), and a novel way of measuring the quality of saliency methods. Through experiments and example output, we demonstrate that XRAI is superior to many other methods. We also show that our proposed measurement methods align with visual observations and standard localization metrics.

{\small
\bibliographystyle{ieee_fullname}
\bibliography{attribution}
}
\newpage
\appendix

\onecolumn

\section*{Supplementary Material}
\begin{figure*}[htb]
    \centering
    \includegraphics[width=\textwidth]{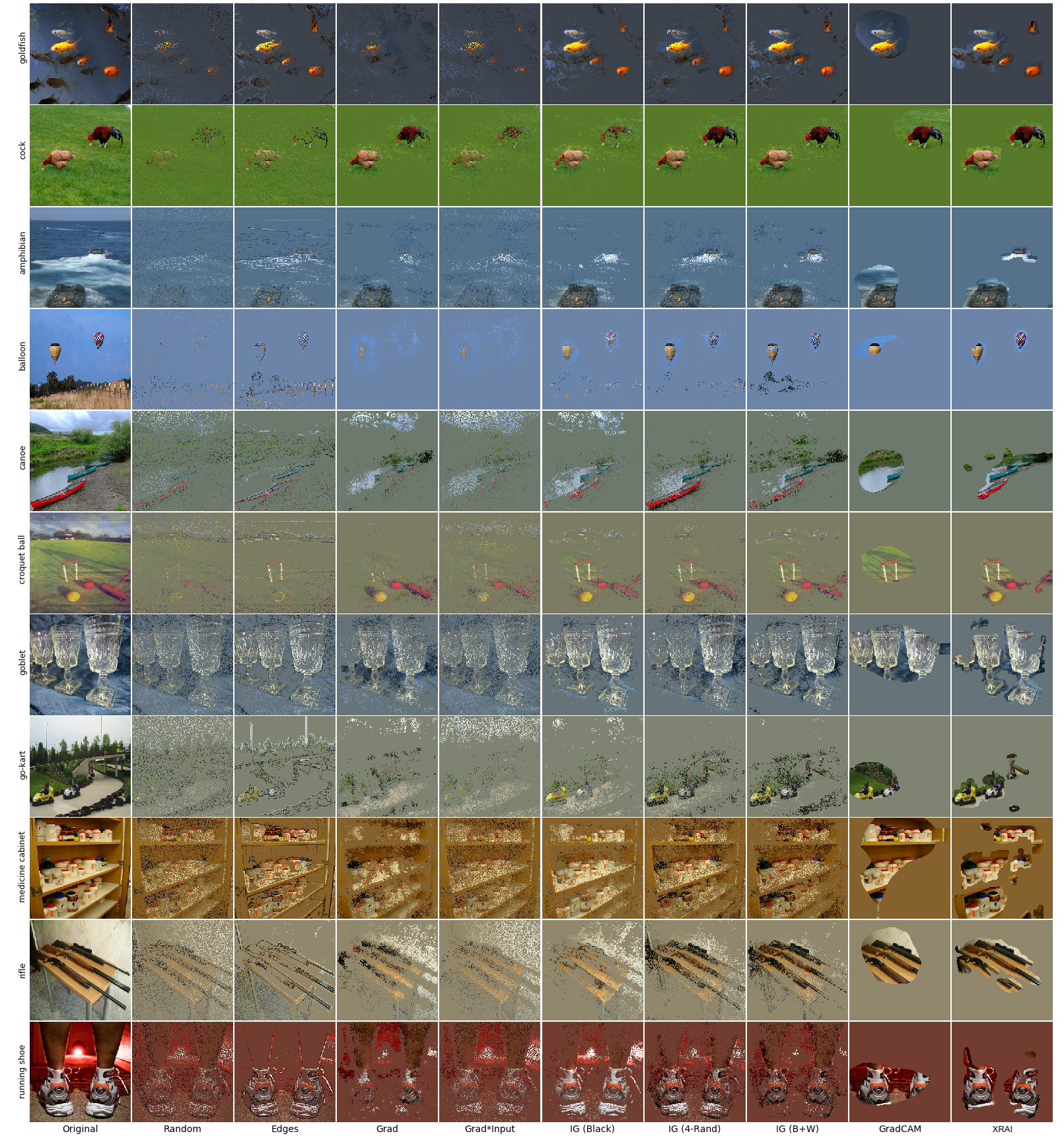}
    \caption{Additional images comparing different saliency methods (as in Figure \ref{fig:methods_comp_vis}).}
    \label{fig:supp_all_methods_1}
\end{figure*}

\begin{figure*}[htb]
    \centering
    \includegraphics[width=0.6\textwidth]{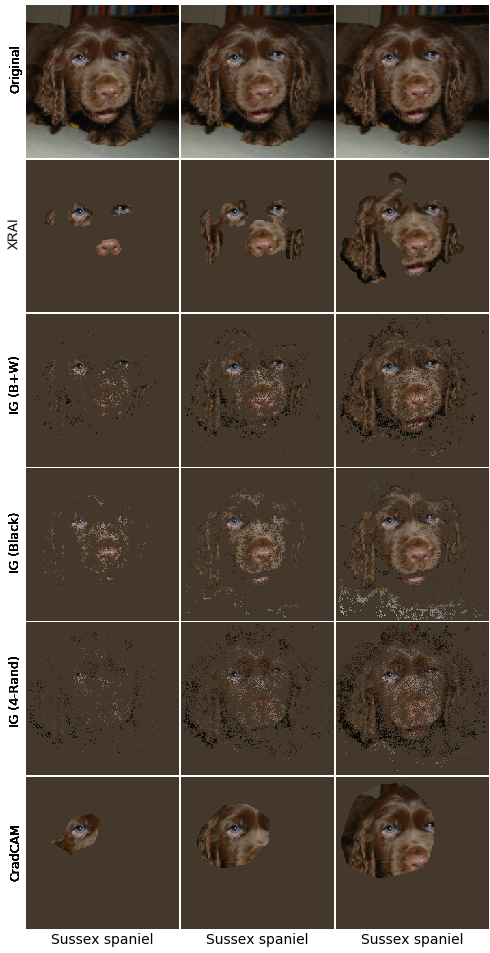}
    \caption{Comparison of methods for an image of Sussex spaniel. GradCAM tends to pick one spot and grow it (one eye of the dog) whereas XRAI can pick both eyes and the snout independently. Ears, eyes and the snout seem to be picked as the most important regions to identify the breed.}
    \label{fig:sup_dog_grow_1}
\end{figure*}

\end{document}